\documentclass{article}

\usepackage{xcolor}

\usepackage{amssymb}

\usepackage{caption}
\usepackage{subcaption}

\usepackage{graphicx}
\usepackage{amsmath}

\usepackage{tabularx}
\usepackage{multirow}

\usepackage{placeins} 

\usepackage[final]{corl_2024} 

\usepackage{cleveref}
\Crefname{algorithm}{Alg.}{Algs.}
\Crefname{equation}{Eq.}{Eqs.}
\Crefname{section}{Sec.}{Secs.}
\Crefname{figure}{Fig.}{Figs.}
\Crefname{tabular}{Tab.}{Tabs.}
\Crefname{table}{Tab.}{Tabs.}

\title{Generative Grasp Detection and Estimation with Concept Learning-based Safety Criteria}

\author{
  Al-Harith Farhad\\
  Innovative Factory Solutions\\
  German Research Center for Artificial Intelligence (DFKI)
  Germany\\
  \texttt{al-harith.farhad@dfki.de} \\
  \And
  Khalil Abuibaid\\
  Chair of Machine Tools and Control Systems (WSKL) \\
  Rheinland-Pfälzische Technische Universität (RPTU)
  Germany\\
  \texttt{khalil.abuibaid@rptu.de} \\
  \And
  Christiane Plociennik\\
  Innovative Factory Solutions\\
  German Research Center for Artificial Intelligence (DFKI)
  Germany\\
  \texttt{christiane.plociennik@dfki.de} \\
  \And
  Achim Wagner\\
  Innovative Factory Solutions\\
  German Research Center for Artificial Intelligence (DFKI)
  Germany\\
  \texttt{achim.wagner@dfki.de} \\
  \And
  Martin Ruskowski\\
  Innovative Factory Solutions\\
  German Research Center for Artificial Intelligence (DFKI)
  Germany\\
  \texttt{martin.ruskowski@dfki.de} \\
}

\begin{document}
\maketitle


\begin{abstract}
Neural networks are often regarded as universal equations that can estimate any function. This flexibility, however, comes with the drawback of high complexity, rendering these networks into black box models, which is especially relevant in safety-centric applications.  
To that end, we propose a pipeline for a collaborative robot (Cobot) grasping algorithm that detects relevant tools and generates the optimal grasp.
To increase the transparency and reliability of this approach, we integrate an explainable AI method that provides an explanation for the underlying prediction of a model by extracting the learned features and correlating them to corresponding classes from the input. These concepts are then used as additional criteria to ensure the safe handling of work tools.
In this paper, we show the consistency of this approach  and the criterion for improving the handover position.
This approach was tested in an industrial environment, where a camera system was set up to enable a robot to pick up certain tools and objects.
\end{abstract}



\section{Introduction}
	
Robots in industry are typically programmed to perform a single task, quickly and precisely, with little to no regard to their surroundings at a given moment in time. Nevertheless, in the past, the safety of human workers was achieved through separating human and robotic spheres of action~\cite{KRUGER2009628}. However, with the increased popularity of collaborative robots (Cobots) working alongside humans, safety becomes an even more critical issue, as separation is no longer an option.

One of the primary tasks Cobots are programmed to perform is grasping various objects, which is achieved through a variety of grasp estimation and generative grasping approaches. Some methods simulate multiple grasping positions and sample the most successful one \cite{mahler2017dex,mahler2019learning}, while others tend to estimate the position of the object in the space by constructing a 6 DoF model of it, after which the optimal grasp is decided \cite{Detry2017SceneUnderstaning}. Other methods aim to generate the grasp position by using Convolutional Neural Networks (CNNs) to generate filters such as quality, width and angle of grasp~\cite{Morrison2018ClosingTL}. 

On the other hand, a lot of effort has been put into the field of Explainable Artificial Intelligence (XAI). In particular, efforts in computer vision, encompassing occlusion-based, gradient-based techniques such as feature attribution, and Shapley value-based techniques~\cite{Lime,Selvaraju2016GradCAMVE,zhou2016Localization}, employed the analysis of the trained model to identify the main features within an image that contributed the most to the prediction. While valuable, this separation between explanation and prediction of post-hoc approaches poses its own set of challenges. When explanations fall short, determining whether the fault lies with the explanation method itself or if the model merely relied on superficial correlations for its predictions can be tricky~\cite{Rudin2018StopEB}. Hence, there is a growing demand for approaches that integrate explanation and prediction within an ante-hoc framework, enabling models to inherently learn explanations.

In this paper, we propose the integration of concept learning, an ante-hoc explainability approach that extracts features from the trained model using the object detection and a generative grasping approach. Two datasets are also created to train each of these components. This serves as a pipeline for detecting relevant tools on the work surface, generating grasps for each, while incorporating an additional logic-based criterion for safe handling of hazardous objects in the workers' environment.

The rest of the paper is structured as follows: In \cref{sec:related}, relevant literature and related work are discussed, then the methodology and dataset used for our approach are introduced in \cref{sec:methodology}. The set-up and results of our experiments are shown in \cref{sec:experiments}, which are discussed along with the limitations of our approach in \cref{sec:discussion}. The paper ends with the conclusion and future work in \cref{sec:conclusion}.

\section{Related Work}
\label{sec:related}

The main goal of this paper is the development of a pipeline that allows for the detection and grasping of work tools with an increase in the safety of Cobots by integrating explainable models into the grasping model. Most existing research focuses on the usage of AI methods to ensure the safety of the shared environment between worker and robot as determined by a review on human-robot collaboration \cite{matheson2019human}. It highlights that numerous studies focus on developing techniques to predict and avert collisions, which is essential for the safe integration of Cobots in a manufacturing setting. Our approach focuses on the safety of the grasp of a parallel gripper during handover protocol to a human worker. For that, the relevant literature is concerned with grasp estimation methods as well as XAI methods.

\subsection{Grasp Estimation and Sampling}
Task-agnostic robotic grasping is a prevalent challenge that encompasses both perception and control. A robotic grasping system is generally composed of the following sub-systems: (a) the grasp detection system, (b) the grasp planning system, and (c) the control system \cite{du_vision-based_2021}, as shown later in \cref{sec:setup}. Among them, the grasp detection system is the key entry point. The majority of existing methods are concerned with parallel grippers \cite{mahler2017dex} due to their relatively low cost and availability. However, more complex grasping tasks can only be achieved with dexterous grippers~\cite{du_vision-based_2021}. These methods require more detailed input such as high-DoF inputs of the object~\cite{Liu2019GeneratingGP} or other multi-dimensional surface fitting approaches \cite{fan2019efficient}.

In general, grasping algorithms can be categorized as follows: (a) imitation-based methods, (b) sampling-based methods, and (c) end-to-end learners \cite{zhang2022robotic}. Imitation-based methods use a dataset of stable grasps and their corresponding objects and attempt to transfer these grasps to similar objects based on the similarity between objects and robot configuration~\cite{Bonardi2019LearningOI,coninck2019learningGrasp}. Sampling-based methods sample a set of possible candidates, among which a discriminator is used to find the best one. Finally, end-to-end methods use a single neural network, that takes an input of raw observations to output the proper grasp configurations, an example of this is the Oriented Anchor Box mechanism (OAB)~\cite{Zhang2019Anchor}.

In this work, we opt to use an end-to-end CNN-based generative grasping approach that was introduced in \cite{Morrison2018ClosingTL}, the approach will be elaborated further in \cref{subsec:ggcnn}.

\subsection{Concept Learning}
\label{subsec:concept}
A review of modern explainable AI taxonomies concluded that all taxonomies classify explainability into either ante-hoc or post-hoc methods \cite{TaxTimo2022}. 
The difference between these two approaches lies in their implementation: ante-hoc methods involve training models such that they become inherently explainable, while post-hoc approaches aim to interpret black-box models after they have been trained. Additionally, inherently interpretable models, such as decision trees and Bayesian learners, are examples of models that provide transparent explanations by design.

Many of the methods proposed either function as post-hoc approaches on a trained model or involve learning an inherently interpretable model. For instance, quantitative Testing with Concept Activation Vectors (TCAV) \cite{kim2018interpretability} uses directional derivatives with intermediate model features to measure the importance of a user-defined concept in the final model predictions. Although this method does not require complete concept annotations, the explanations are based on prior knowledge of concepts regarding the data points. Another method introduced in \cite{Zhou_2018_ECCV} decomposes model predictions into concept vector projections, utilizing model-generated saliency by CAM \cite{zhou2016Localization}. Explanations can also be provided in the form of important features or inputs to the models. Such is the case with Shapley values \cite{NEURIPS2020_ecb287ff} or other methods such as \cite{NEURIPS2019_77d2afcb}, which works by providing the most important input data features that contribute to a prediction.


\section{Methodology}
\label{sec:methodology}
In this section we present the methods, tools and datasets used to develop this work. Each component of the operational pipeline for the grasping model is explained as well as the additional concept-based safety criteria.

\subsection{Datasets}
\label{subsec:data}
Due to the cumulative nature of the proposed architecture, different datasets were used to train and develop each part of the algorithm as follows:

\textbf{Work tool Dataset:} Since the main objective of the Cobot is the safe handover of common work tool, a dataset of unordered work tools was created in the DFKI lab. The images were taken from 2 different perspectives, top- and side-view. Each image contains one or more work tool and was annotated using the YOLOv5 format which includes the class id, and the 4 coordinates of the bounding box \cite{YOLO}. The images here are RGB images taken at the workbench of the Cobot.

Furthermore, the tools within the bounding boxes were extracted to create a dataset of images containing a single work tool along with its class id. This will be used for the testing and development of the Concept Learning approach.

\textbf{Jacquard Dataset:} The Jacquard Dataset is a dataset of RGB-D images containing a variety of common objects with annotations of possible grips using the format $(x, y, h, w, \theta)$~\cite{Jacquard}, where $x$ and $y$ are the coordinates, $h$, $w$, and $\theta$ are height, width and angle, respectively. The dataset was systematically generated by simulating a scene and attempting different grip positions and gripper sizes. It is also reported to produce better results than similar datasets.

Additionally, a subset of the Work tool dataset was similarly annotated to provide a dataset that could be potentially used to fine-tune the trained grip estimation model to the specific usecase and work tools in the environment at hand.

\subsection{Operational Pipeline}
\begin{figure}[t!]
    \centering
    \includegraphics[width=0.75\textwidth]{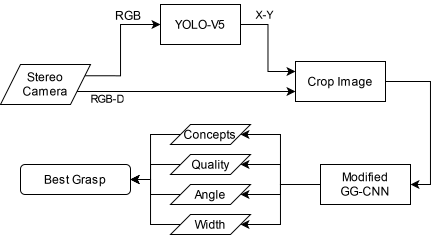}
    \caption{Grasp Estimation Pipeline}
    \label{fig:pipeline}
\end{figure}

The proposed pipeline has the architecture shown in \cref{fig:pipeline}. The pipeline starts with a stereo-depth camera. An RGB image is fed into an object detection neural network, YOLOv5, which is trained to detect work tools as mentioned in \cref{subsec:data}. The detected objects are extracted and fed into the modified GG-CNN as an RGB-D image, where the best grasp is estimated using the output of the network, which will be explained in further detail in \cref{subsec:ggcnn}.

\subsubsection{Object Detection}

YOLOv5 (You Only Look Once version 5) is a prominent member of the YOLO family of object detection (OD) algorithms \cite{YOLO}, which was one of the first methods, along with SSD (Single-Shot MultiBox Detector) \cite{ssd} to perform detection in an image through a single forward pass. 


The architecture of YOLOv5 is designed to optimize speed and accuracy, employing a combination of advanced techniques which enable it to achieve superior results on various benchmark datasets, making it a good candidate for our work. Due to the hardware limitations of the setup, YOLOv5 was chosen as the primary algorithm, due to its relatively lower complexity, compared to its newer counterparts: v6, v7 and v8. However, a comparison to YOLOv8 was also done in \cref{sec:results} to validate the choice.

\subsubsection{Generative Grasping CNN}
\label{subsec:ggcnn}

The Generative Grasping Convolutional Neural Network (GG-CNN) utilizes deep learning techniques to predict grasp configurations directly from visual input \cite{Morrison2018ClosingTL}. Unlike traditional methods, 
GG-CNN streamlines the task by generating a dense grasp map in a single pass. This design allows the network to evaluate potential grasps across the entire image efficiently, making it well-suited for real-time applications.

The GG-CNN architecture is fully convolutional, enabling it to generalize well to various objects. 
The network is trained to predict the quality and configuration (angle and width) of grasps on a pixel-wise basis, which allows it to handle a wide range of object shapes and sizes with great accuracy.

\subsubsection{Grasping Safety Criteria}
A safety criterion is developed by learning the concepts associated with each detected tool. The presence of a certain concept, or lack thereof, triggers the filtration and some grip positions, and allows the handover to the worker. 

As mentioned in \cref{subsec:concept}, Network Dissection is one of the most exhaustive methods in the XAI field. However, due to the multi-layer evaluation of its features and the amount of data needed for evaluation, it cannot be used effectively in a real-time setting where the Cobot needs to make relatively swift decisions.

Therefore, our proposed approach adds an additional layer to the grasping model which acts as a concept layer, a quasi feature extractor that can decipher concepts from the given images. This is done by correlating each extracted feature to its corresponding output class, providing an added advantage of extracting ante-hoc explanations, as well as simpler computation at run time, as the majority of the correlation can be computed during model development.

\begin{figure}[h!]
    \centering
    \includegraphics[width=0.7\textwidth]{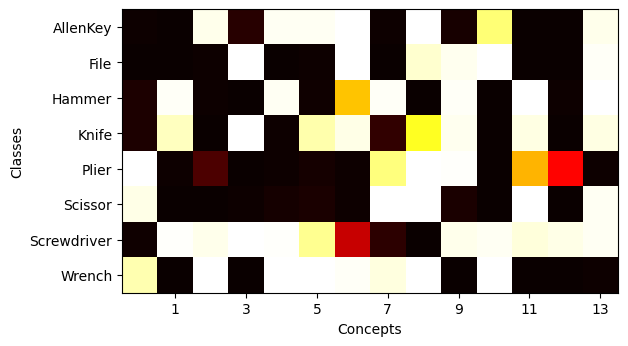}
    \caption{Feature-Class Correlation Heatmap}
    \label{fig:heatmap}
\end{figure}

By correlating these features with their corresponding classes, the grasp can be fine tuned to focus on a specific part of the tool to hold it safely. This is shown in \cref{fig:heatmap} where a heatmap of features and classes shows that certain features would only be triggered when a certain tool is being grasped. Initial inspection shows a clear correlation between some of the triggered features for classes such as the Pliers, Scissors, and Wrench, which are have similar symmetries. The file, knife, and screwdriver also have similar activations and are indeed similar looking.


\section{Experimental Evaluation}
\label{sec:experiments}

To validate the method proposed, a universal robot arm was used with RGB-D stereo cameras. The exact setup and architecture is explained in \cref{sec:setup}. The experimental results are also presented in \cref{sec:results} for each component of this approach.

\subsection{Setup}
\label{sec:setup}
\begin{figure}[h!]
    \centering
    \includegraphics[width=0.8\textwidth]{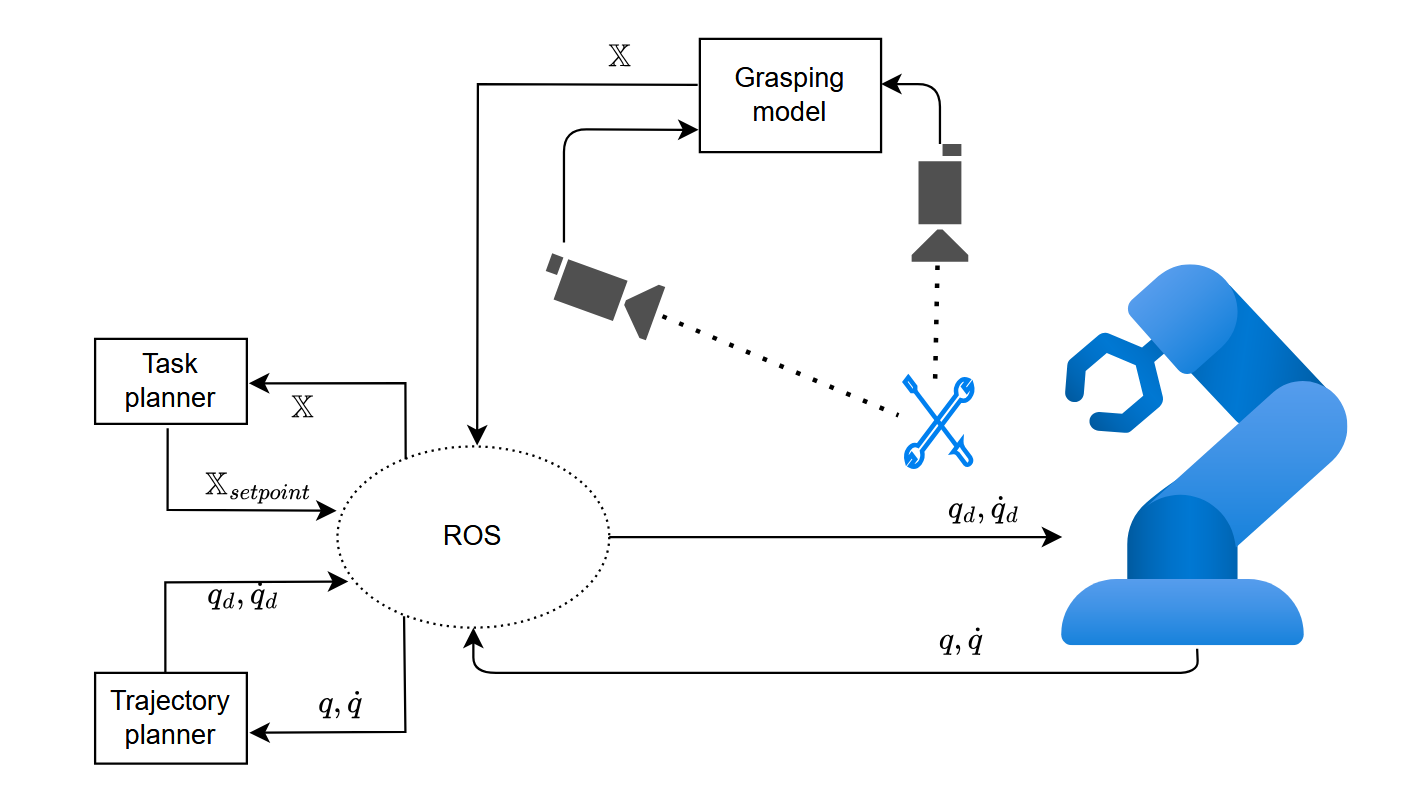}
    \caption{Control and Communication Architecture of the Cobot}
    \label{fig:ctrlart}
    \FloatBarrier
\end{figure}

In the experimental setup, a UR5e robotic arm, renowned for its flexibility and precision, is 
integrated with a dual ZED2i stereo camera system from Stereolabs via the Robot Operating System (ROS) Noetic to facilitate precise object detection, safe grasping operations, and a safe path plan to hand over the object to a human operator. The control architecture is depicted in \cref{fig:ctrlart}. The ROS Noetic communication framework enables seamless data exchange and real-time coordination between the robotic arm and the cameras, ensuring efficient and accurate operations. 

Comprehensive 3D visual data of the target object from multiple angles are captured by the two strategically positioned ZED2i cameras. These high-resolution images are collected for the grasping model for further analysis.

The grasping model processes the visual data to identify the object's type, dimensions, and exact location within the workspace. This information is compiled into a dataset, denoted as $\mathbb{X} = (x, y, h, w, \theta)$, which is then transmitted to the task planner. Upon receiving $\mathbb{X}$ via ROS, the task planner generates setpoints, $\mathbb{X}_{setpoint}$ that define the precise coordinates and actions the UR5e should execute, including the optimal grasping angle and path it should take between setpoints.

The trajectory planner plays a crucial role by utilizing sophisticated path planning algorithms to calculate an optimal trajectory between the setpoints. These algorithms ensure that the robotic arm moves smoothly and efficiently by sending signals to the UR5e driver via ROS a desired joint position and velocity which are denoted by $q_{d}$ and $\dot{q}_{d}$ respectively. The scheme is illustrated in fig \cref{fig:ctrlart}. In response, the system receives the robot's joint position and velocity readings, forming a closed-loop structure that guarantees the optimal path.

Additionally, two PCs are used to run the process. One that holds the ROS-master as well as running the UR5e driver. It is running on Ubuntu 20.04 LTS real-time kernel with an i7 processor. The second PC is connected to two ZED2i stereo cameras and runs the grasping model with Nvidia GeForce GTX 1070 as GPU and Ubuntu 22.04 LTS as operating system. The network setup are done via Ethernet on all devices. The experiment involves the application of the proposed algorithm pipeline on multiple tools using this setup.

\subsection{Results}
\label{sec:results}
\begin{figure}[t!]
    \centering
    \includegraphics[width=0.92\textwidth]{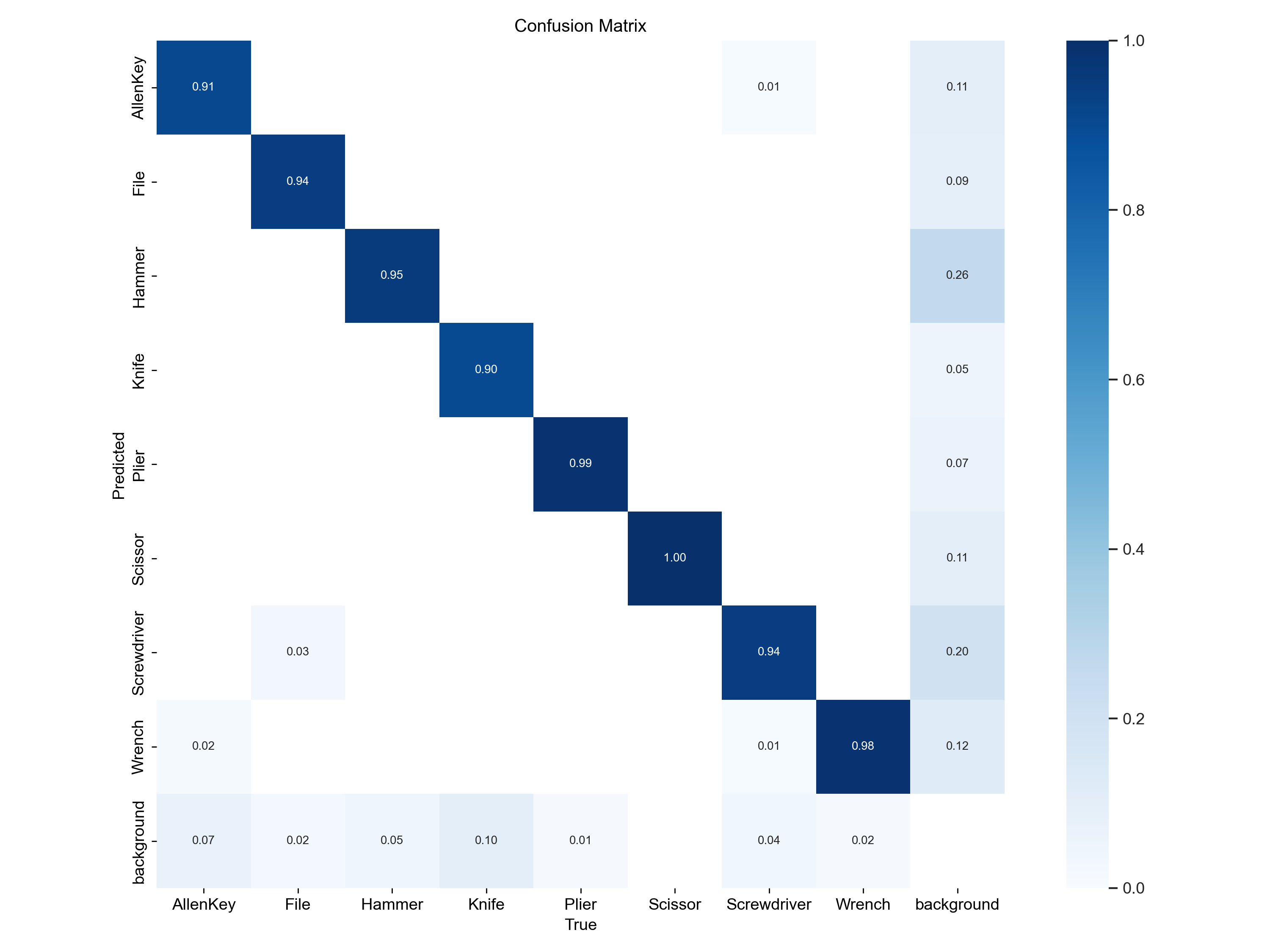}
    \caption{Confusion Matrix of the YOLOv5 for Tool Detection}
    \label{fig:yolocm}
    \FloatBarrier
\end{figure}

The YOLOv5 network trained for the work tool detection managed to achieve a good performance when trained on the dataset of real tools. However, as shown in \cref{fig:yolocm}, the model did have a few misclassifications, especially between visually similar objects such as the Screwdriver and File. Additionally, due to the silver background of the work surface, mostly metallic tools with high reflectivity were particularly at a disadvantage such as the knife and Allenkey.

Furthermore, training a more complex algorithm such as YOLOv8 did not result in a significant improvement in detection, in comparison to YOLOv5. This can be attributed to the relative homogeneity of the dataset as it only contains 8 tools, without much variation in shape. The performance comparison is shown in \cref{tab:yolov5v8}. Both networks were trained with the same parameters.

\begin{table}[h]
    \centering
    \resizebox{\textwidth}{!}{%
    \begin{tabular}{|c|c|c|c|c|c|c|c|c|c|c|}\hline
         & Measure & All &Allenkey & Hammer & File & Knife & Plier & Scissor & Screwdriver & Wrench \\\hline
         \multirow{2}{*}{YOLOv5}& mAP50 & \textbf{0.96} & \textbf{0.913} &  \textbf{0.967} & 0.958 & \textbf{0.921} & \textbf{0.993} & \textbf{0.993} & 0.946 & \textbf{0.986}\\\cline{2-11}
         & mAP50-95 & 0.74 & 0.692 & 0.783 & 0.777 & \textbf{0.693} & \textbf{0.749} & 0.794 & 0.691 & 0.744\\\hline
         \multirow{2}{*}{YOLOv8}& mAP50 & 0.945 & 0.899 & 0.961 & \textbf{0.977} & 0.853 & 0.993 & 0.935 & \textbf{0.953} & 0.985 \\\cline{2-11}
         & mAP50-95 & \textbf{0.775} & \textbf{0.714} & \textbf{0.801} & \textbf{0.837} & 0.658 & 0.714 & \textbf{0.807} & \textbf{0.77} & \textbf{0.803}\\\hline
    \end{tabular}%
    }
    \caption{mAP of YOLOv5 and YOLOv8}
    \label{tab:yolov5v8}
\end{table}

On the other hand, training the modified CNN model with the concept layer showed a clear presence of prominent features, as shown in \cref{fig:heatmap}. These features were correlated with concepts defined to fine-tune the grip such that it would be rotated for a safer configuration when handing it over to the worker.

The performance of the GG-CNN however was measured experimentally by attempting to pick-up objects off the work surface. In this experiment we excluded any failed attempts due to the failure of the object detection algorithm and managed to achieve an overall success rate of 81.4\%, out of a total of 70 of attempts.

    \begin{table}[h]
        \centering
        \begin{tabular}{|c|c|c|c|c|}
             \hline
             & GG-CNN\cite{Morrison2018ClosingTL} & Dex\cite{mahler2017dex} & OAB\cite{Zhang2019Anchor} &  Ours \\\hline
            Simple Objects & 91\% & 93\% & 90\% & -\\\hline
            Complex Geometry & 81\% & 83\% & 84.2\% & 81.4\%\\\hline
        \end{tabular}
        \caption{Comparison to other similar approaches}
        \label{tab:comparasion}
    \end{table}

The success rate achieved is comparable to other methods in this field as shown in \cref{tab:comparasion}. Due to the different datasets used in each paper, we grouped common objects such as household objects and cylindrical objects into simple, while more objects with more complicated or adversarial geometries were treated as complex.

\begin{figure}[htb!]
\minipage{0.32\textwidth}
  \includegraphics[width=\linewidth]{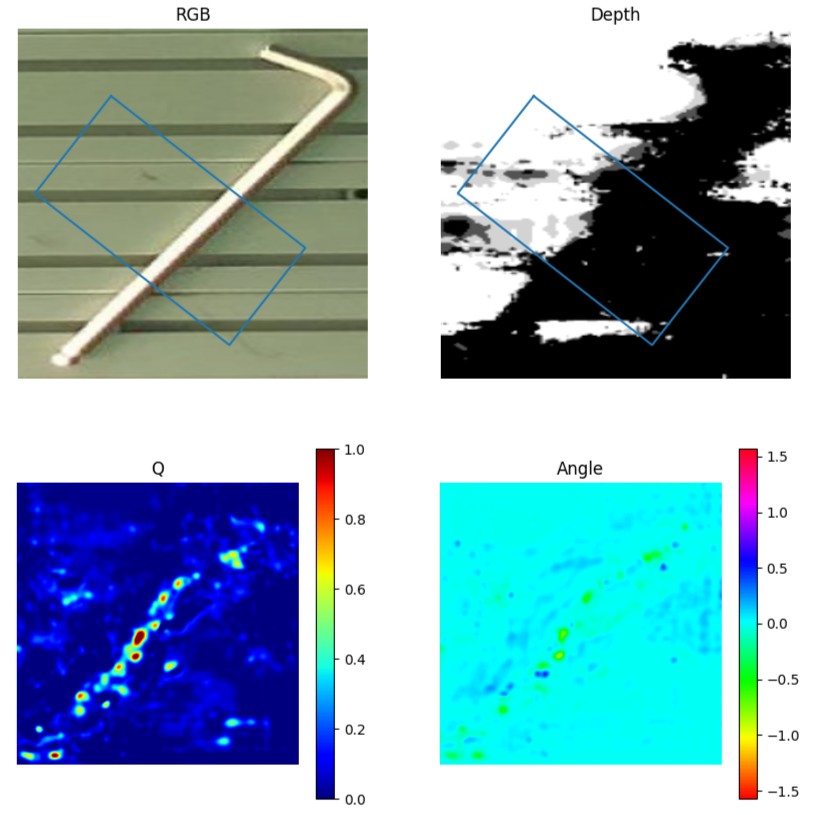}
  \caption{Successful Grasp Attempt}\label{fig:awesome_image1}
\endminipage\hfill
\minipage{0.32\textwidth}
  \includegraphics[width=\linewidth]{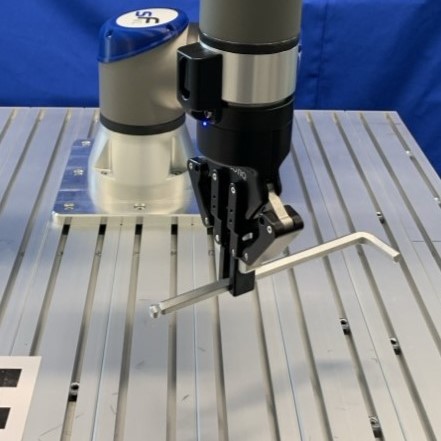}
  \caption{Cobot Grasping Tool Successfully}\label{fig:awesome_image2}
\endminipage\hfill
\minipage{0.32\textwidth}%
  \includegraphics[width=\linewidth]{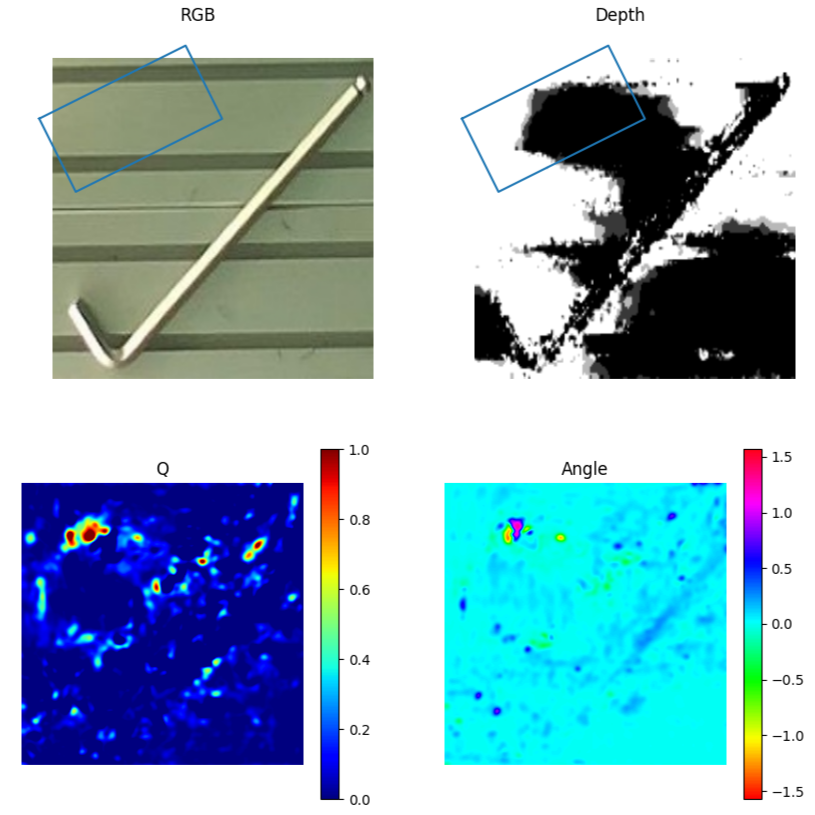}
  \caption{Failed Grasp Attempt}\label{fig:awesome_image3}
\endminipage
\end{figure}


\section{Discussion and Limitations}
\label{sec:discussion}
\subsection{Discussion}
As mentioned in \cref{sec:related}, parallel gripper grasp estimation algorithms are some of the most researched algorithms in the field. However, the systematic pipeline combining multiple steps for the detection and grasping of work tools, as presented in this paper, offers an improvement to its individual components in that it is selective with the tools being grasped, even when presented with a variety of objects.

Although the model for tool detection had a few misclassifications between tools that are similar in appearance, such as Knife and File, these misclassifications did not directly impact the grasping algorithm. For instance, if tools like the already mentioned Knife and File were misclassified, the wrong class did not affect the grasping algorithm as it attempted to generate a grasp nevertheless. The size of the bounding box is more important, as it will feed that information directly to the grasp estimation algorithm.

Furthermore, an explainablity component was also integrated into the convolutional neural network that works as a key feature extractor for the concepts learned by the CNN. This concept layer can also be implemented on a trained model by freezing the original weights and updating only the weights necessary for activation. This leaves the performance of the model seemingly unaffected.

\subsection{Limitations}



From an execution point of view, several challenges need to be addressed to improve the system's performance, such as lighting conditions which play a critical role, as variations in illumination can significantly affect the accuracy of the grasping model. This can be improved by integrating a lighting fixture which will result in consistent lighting conditions. Alternatively, creating more variation within the training dataset and applying certain filters to the images might also lead to improvement in the system.

Furthermore, grasping an object from the table requires precise movements due to the object's height relative to the table surface. Any inaccuracy in this process may cause the robot to collide with the table, risking damage to the robotic arm. However, this limitation is imposed by the available hardware and can only be resolved by using more advanced equipment or through the implementation of sensor fusion methods.

From an algorithmic point of view, due to the integration of multiple AI models, a failure of one might cause a chain of cascading failures. A causality study is an open question in academia, and may provide useful insights in the future. However, in the current system design, incorrect bounding boxes or missed objects are more likely to cause catastrophic failures than false predictions or misclassifications. Moreover, adding safeguards between the components may improve the performance.


\section{Conclusion}
\label{sec:conclusion}

In this paper, a pipeline and control architecture was introduced for the safe handover of work tools using a collaborative robot (Cobot). A YOLOv5 network was trained to detect work tools, whose location is then fed into a generative grasping convolutional neural network GG-CNN, which is an end-to-end algorithm that directly determines the best grasp, by predicting the width, angle and quality of the grasp.

Using this approach, the Cobot managed to achieve a success rate of 81.4\%, which is comparable to the results obtained by existing approaches, however with the added precision of only grasping work tools, and providing an additional explainablity aspect.

The CNN was modified to add a feature extraction layer. The extracted features were correlated to their corresponding output classes. These concepts were used as refinement criteria for the grasping position, based on a defined criterion for the concepts to move the tool and hand it over safely. Another advantage of the concept layer is that it can easily be integrated into trained networks, requiring only simple fine-tuning without affecting the overall performance of the model.

This approach is hampered by a few limitations, such as the lack of clearly defined path planning and force constraints, the dependency on lighting conditions, as well as the chain effect of the cascading models. An investigation of these points could prove valuable, leading to increased safety and overall system efficiency.




\bibliography{references}  

\end{document}